\documentclass[conference]{IEEEtran}
\usepackage{spconf,amsmath,graphicx,hyperref}
\usepackage{amssymb}
\usepackage{subcaption}
\usepackage{xspace}
\usepackage{bm}
\newcommand\da{DA\xspace}
\newcommand\unet{U-Net\xspace}

\usepackage[utf8x]{inputenc}

\title{SODA: Self-organizing data augmentation in deep neural networks - Application to biomedical image segmentation tasks\\
\thanks{\copyright 2022 IEEE.  Personal use of this material is permitted.  Permission from IEEE must be obtained for all other uses, in any current or future media, including reprinting/republishing this material for advertising or promotional purposes, creating new collective works, for resale or redistribution to servers or lists, or reuse of any copyrighted component of this work in other works.}
}

\name{Arnaud Deleruyelle, John Klein, Cristian Versari}
\address{Univ. Lille, CNRS, Centrale Lille, UMR 9189 - CRIStAL, F-59000 Lille}
\IEEEoverridecommandlockouts 
\begin{document}
%
\maketitle
\begin{abstract}
In practice, data augmentation is assigned a predefined budget in terms of newly created samples per epoch. When using several types of data augmentation, the budget is usually uniformly distributed over the set of augmentations but one can wonder if this budget should not be allocated to each type in a more efficient way. 
This paper leverages online learning to allocate on the fly this budget as part of neural network training. This meta-algorithm can be run at almost no extra cost as it exploits gradient based signals to determine which type of data augmentation should be preferred. Experiments suggest that this strategy can save computation time and thus goes in the way of greener machine learning practices. 
\end{abstract}
\section{Introduction}
\label{sec:intro}
The benefits of data augmentation (DA) to train deep neural networks has been widely acknowledged. This technique consists in applying various transformations to training examples in order to obtain new ones. Provided that the data distribution is invariant with respect to these transformations, the newly created samples help learning a model with better generalization performances because one is learning from a larger training set. Even when the distribution is not invariant with respect to the transformations, \da can also be beneficial and prevent from overfitting the original training dataset.

There are many possibilities for machine learning practitioners to generate augmented data. We focus in this paper on deep neural network models whose inputs are images. The most popular \da transformations for images include rotations, flipping, noise injection, color or illumination changes, or non-rigid deformations among others. More recently, there has been attempts to use previously trained generative models to sample new data \cite{zhu2018emotion,synthDataAugFrid2018}. Such models can also create new samples through style transfer \cite{jackson2019style}. Another line of thought is to learn how to augment in the spirit of \textit{learning-to-learn} meta-learning algorithms \cite{wang2017effectiveness,inoue2018data}. Such approaches usually involve a pair of networks (one to solve the main task and one to learn to augment data meaningfully). Note that data augmentations can be combined, which makes the number of possible choices very large. 
For a review of data augmentation techniques, the reader is referred to \cite{shorten2019survey}. 

Regardless of the nature of data augmentation types, in practice, one has to (i) choose a limited number of them and (ii) define a budget in terms of newly created samples through augmentation. Sampling too many augmented data might be counter-productive in terms of exploitation of computational resources. In general, it is however hard to anticipate which \da type should be chosen. In addition, the appeal of a given \da type might not be constant across training epochs, i.e. the problem is not a mere scalar hyperparameter setting issue but involves determining a whole sequence of actions.

This paper investigates a mean of learning on the fly which \da types should be preferred through already computed gradients as part of neural network training through gradient based optimizers. Because we exploit gradient signals, our approach has almost no extra cost compared to backpropagation. Building upon HEDGE, an online learning algorithm, we design an algorithm that adaptively determines a score for each \da type and reduces/increases the number of samples obtained from each of them for the next epoch accordingly.

The closest related work with respect to our contribution is the AutoAugment approach from Cubuk et al. \cite{Cubuk_2019_CVPR}. This approach uses reinforcement learning (RL) to find an efficient sequence of \da types and magnitudes for each batch. The RL procedure requires to re-train a neural network and evaluate its accuracy in order to obtain a feedback that allows converging to a meaningful policy. This is in sharp contrast with the proposed approach in which the neural network is trained only once and the sequence of \da types is determined on the fly. 
While the number of possible sequences can be extremely large, the problem can still be regarded as a hyperparameter optimization one. But again, techniques such as Bayesian Optimization or Multi-Armed Bandits (MABs) \cite{huang2020asymptotically} from the literature require to evaluate a sequence through repeated training runs of the chosen neural network architecture. 

The paper is organized as follows. The next section formalizes the \da allocation problem and presents the HEDGE framework. 
Section \ref{sec:a_new_self_organized_data_augmentation} provides a detailed presentation of our contribution whose backbone is HEDGE. 
The proposed algorithm is evaluated on \unet \cite{ronneberger2015u} architectures for segmentation tasks on biomedical images. Experimental material is presented in section \ref{sec:experimental_results} and suggests that our approach allows saving computation time compared to naive policies while achieving comparable accuracy.


\section{Problem statement and background} 
\label{sec:problem_statement_and_background}

\subsection{Preliminaries} 
\label{sub:preliminaries}


In the supervised learning setting, one has access to a training dataset $\mathcal{D}$ which contains $n$ pairs $\left( \mathbf{x}, y \right)  $ of inputs/targets. Given a parametric model such as a neural network $f_{\boldsymbol\theta}$, our task consists in determining the best vector of trainable parameters $\boldsymbol\theta$ such that, for each training example, $L \left( f_{\boldsymbol\theta} \left( \mathbf{x} \right) ,y \right) $ is small where $L$ is the chosen loss function.
If $n$ is large enough, the training algorithm issuing the model minimizes the expected loss (in a probably approximately correct sense).

Now, suppose one can afford to learn from $n_a$ additional augmented training examples per epoch and has $K$ \da generators to choose from to allocate this computational budget. 
Given a $K$ dimensional vector $\boldsymbol\pi$ of probabilities, a $\pi_k$ fraction of the $n_a$ augmented examples is queried to the \da generator $k$. In the current paper, we are interested in estimating a meaningful vector $\boldsymbol\pi$. Because the appropriate proportion of each type of \da is not constant over training epochs, we will use an index $t$ for this vector in the sequel. Moreover, we assume $n_a \gg K$ so that each \da receives at least a budget of one point.

\subsection{Online learning with HEDGE} 
\label{sub:online_learning_with_hedge}


Let us further assume that the neural network training environment provides a $K$-dimensional bounded action-loss signal $\bm{\ell}_t$ that tells us how much each \da type is unhelpful in training our model. This setting is an online optimization problem of learning with expert advice where the learner suffers a loss given by the following dot product $\boldsymbol\pi_t^\intercal \cdot \bm{\ell}_t$ (at each time step). 
HEDGE \cite{freund1997decision2} is an algorithm achieving minimal regret w.r.t. the cumulative loss $\sum_{t=1}^T \boldsymbol\pi_t^\intercal \cdot \bm{\ell}_t$ \cite{cesa2006prediction} where $T$ is the final epoch (assumed to be fixed here for simplicity).

HEDGE is also known as the aggregating algorithm or exponentially weighted forecaster. 
Unlike MABs which receive a feedback only for the previously chosen action, HEDGE receives\textbf{} a feedback for all actions. 
In our context, at each epoch $t$, HEDGE consists in the following sequence of steps:
\begin{enumerate}
  \item Learner chooses $\boldsymbol\pi_t$, where $\pi_{k,t} = \frac{w_{k,t}}{\sum_{k=1}^K w_{k,t}}$.
  \item Environment reveals $\bm{\ell}_t$.
  \item Learner updates weights $w_{k+1,t} = w_{k,t} \exp \left( - \eta \ell_{k,t} \right)$ for each $k$.
\end{enumerate}

In the last step, $\eta$ is HEDGE learning rate. The weights characterize how helpful each \da type is. Before the first epoch, they are initialized as $w_{k,1}=1, \forall k$. HEDGE can be employed in a context where the environment is adversarial (and chooses $\bm{\ell}_t$ to maximize regret). In this paper, the environment does not exhibit such a behavior, but may evolve with respect to $t$ and thus we need to elaborate on this algorithmic procedure to fulfill our goal, as explained in the next section.
It should be underlined that HEDGE has the advantage of not requiring action-loss distribution assumptions.


\section{A new self-organized data augmentation allocation algorithm} 
\label{sec:a_new_self_organized_data_augmentation}

This section introduces a \da allocation algorithm that can be coupled with neural network optimization at very small extra-cost. HEDGE is the backbone of this algorithm. 
Building upon HEDGE, we provide a workable and contextualized algorithm for \da allocation deployed in only one training run, in contrast with typical MAB based or offline allocations which use multiple runs.

\subsection{Crafting a meaningful feedback signal} 
\label{sub:crafting_a_meaningful_feedback_signal}
As part of HEDGE, the environment (the neural network training procedure in our case) should reveal at each epoch $t$ how each option (\da type in our case) is unhelpful through loss values. In our context, and among other possibilities, the $k^{\text{th}}$ action-loss is defined as the average train-loss discrepancy obtained by learning from this type of \da:
 
\begin{equation}
  \frac{1}{n_{k,t}} \sum_{i=1}^{n_{k,t} } \frac{1}{n} \sum_{\left( \mathbf{x},y \right)\in \mathcal{D}  }L \left( f_{\boldsymbol\theta + \Delta\boldsymbol\theta_{t,i}  } \left( \mathbf{x} \right) ,y \right) - L \left( f_{\boldsymbol\theta } \left( \mathbf{x} \right) ,y \right),\label{eq:loss_diff}
\end{equation}
where $n_{k,t}$ is the number of examples allocated to the $k^{\text{th}}$ \da at epoch $t$ and $\Delta\boldsymbol\theta_{t,i} $ is the parameter update obtained from the $i^{\text{th}}$ augmented training example in this \da category. Note that $n_{k,t}\geq 1$ is a positive integer obtained by applying a rounding function to $\pi_{k,t}\times n_a $ so that $\sum_k n_{k,t} = n_a$.

It would be computationally prohibitive to evaluate the train-loss discrepancy for each augmented training example ($n$ forward passes per augmented sample). To circumvent this issue, let us view the train-loss as a function $J$ of $\boldsymbol\theta$. Using first order Taylor expansion, we have $J \left( \boldsymbol\theta + \Delta\boldsymbol\theta_{t,i}  \right) \approx J \left( \boldsymbol\theta \right) + \nabla J \left( \boldsymbol\theta \right)^{\intercal}\cdot  \Delta\boldsymbol\theta_{t,i}$. Moreover, assuming the neural network is optimized by stochastic gradient descent (SGD), then $\Delta\boldsymbol\theta_{t,i} = - \alpha \mathbf{g}_{t,i}^{(k)} $ where $\mathbf{g}_{t,i}^{(k)}$ is the gradient computed from the augmented training example and $\alpha$ is SGD learning rate. 
Consequently, the train-loss discrepancy can be approximated by a dot product of gradients. 

The gradient $\nabla J \left( \boldsymbol\theta \right)$ remains prohibitive to compute. But, alleging that during one epoch the network will not evolve too much\footnote{This holds for a sufficiently small $\alpha$ which is also required for the Taylor approximation to be good.}, this gradient can be replaced by the average gradient $\mathbf{g}_{t}^{(0)}$ obtained from data points in $\mathcal{D}$ throughout SGD execution. This is actually the idea behind SGD. 
Consequently, $- \alpha \left.\mathbf{g}_{t}^{(0)}\right.^{\intercal} \cdot \mathbf{g}_{t}^{(k)}$ can be used as proxy for \eqref{eq:loss_diff} where $\mathbf{g}_{t}^{(k)}$ is the average gradient obtained from augmented data points from the $k^{\text{th}}$ \da throughout SGD execution in the current epoch. 
Finally, to comply with HEDGE boundedness requirements, we will use the following normalization based on the cosine of the angle between the gradients (which also removes constant $\alpha$):

\begin{equation}
  \ell_{k,t} = \frac{1}{2} \left( 1 - \frac{\left.\mathbf{g}_{t}^{(0)}\right.^{\intercal} \cdot \mathbf{g}_{t}^{(k)} }{\left\lVert \mathbf{g}_{t}^{(0)}\right\rVert_2 \left\lVert \mathbf{g}_{t}^{(k)}\right\rVert_2} \right),\label{eq:loss_approx}
\end{equation}

where $\lVert .\rVert_2$ is the Euclidean $\mathcal{L}_2$ norm. 
 The action-loss \eqref{eq:loss_approx} is easy to interpret: it is a gradient matching based signal which has been also used in offline DA policy allocation learning \cite{DirectDiffAug,zheng2021deep}. 

In order to work with more stable action-loss signals and mitigate gradient direction oscillations, we will use momentum estimates of average gradients, i.e. $\ell_{k,t} = \frac{1}{2} \left( 1 - \frac{\left.\tilde{\mathbf{g}}_{t}^{(0)}\right.^{\intercal} \cdot \tilde{\mathbf{g}}_{t}^{(k)} }{\left\lVert \tilde{\mathbf{g}}_{t}^{(0)}\right\rVert_2 \left\lVert \tilde{\mathbf{g}}_{t}^{(k)}\right\rVert_2} \right)$ where, for any $k \in {0,..,K}$, 

\begin{align}
\mathbf{m}^{(k)}_{t} &= \rho\: \mathbf{m}^{(k)}_{t-1} + \left( 1- \rho \right) \mathbf{g}_t^{(k)},\\
\tilde{\mathbf{g}}_t^{(k)} &= \frac{\mathbf{m}^{(k)}_{t}}{1 - \rho^t},
\end{align}

and $\mathbf{m}^{(k)}_{t}$ is a momentum vector for the $k^{\text{th}}$ \da which is initialized as $\mathbf{m}^{(k)}_{t} = \mathbf{0}$. 





\subsection{Online learning with a discount factor} 
\label{sub:online_learning_with_a_forget_factor}
Now that we have action-loss signals, one could apply HEDGE to solve our \da allocation task. However, as mentioned before, appropriate allocations may vary across epochs. To take into account this time-dependence, we propose to add a discount factor $\beta \in \left[ 0;1 \right] $ in the weight update of HEDGE, that is, step 3. from HEDGE is replaced with 

\begin{equation}
  w_{k+1,t} = w_{k,t}^{\beta} \exp \left(   - \eta \ell_{k,t} \right), \forall k .
\end{equation}
When $\beta=1$, we retrieve HEDGE while when $\beta=0$ we obtain a memoryless procedure. This simple modification of HEDGE allows the \da allocation algorithm to forget about past observed action-losses in order to more rapidly adapt to a shift of \da usefulness over epochs. Unrolling this update rule, a \da weight can be re-written as
\begin{equation}
  w_{k,t} = \exp \left( -\eta \sum_{t'=1}^t \beta^{t-t'}\ell_{k,t'} \right).
\end{equation}
We thus see that this variant of HEDGE is meant to minimize a discounted version of the cumulative loss $\sum_{t=1}^T \beta^{T-t} \boldsymbol\pi_t^\intercal \cdot \bm{\ell}_t$. 

In total, there are three hyperparameters to tune ($\eta, \rho$ and $\beta$) for our \da allocation algorithm which we call SODA for Self-Organizing Data Augmentation.



\section{Experimental validation through biomedical image segmentation } 
\label{sec:experimental_results}

This section provides numerical experiments to evaluate the benefits of SODA. We chose a validation framework that uses \unet architectures \cite{ronneberger2015u} on segmentation tasks. Indeed, in this kind of supervised problems, \da appears to be a critical aspect of the training process so that \unet can generalize efficiently.

\subsection{Datasets \& Preprocessing}

We evaluate our strategy on 4 datasets.
The first dataset is DRIVE \cite{staal2004ridge}. It contains the segmentation of blood vessels in retinal images. 
In the second one \cite{Caps2021}, a transparent capsule crossing in a micro-tube must be segmented.  
The last two datasets are 'DIC-C2DH-HeLa' and 'PhC-C2DH-U373' from the ISBI cell tracking challenge \cite{mavska2014benchmark}.

We only use $n=20$ training examples for each experiment so that \da has a greater impact. 
Inputs are standardized (zero mean and unit variance). 
We use 3 \da types:
\begin{itemize}
\item noise injection: choose $\sigma$ uniformly at random in $\lbrace 0.01, 0.02, \ldots, 0.05 \rbrace$ and multiply the image by a noise whose pixels are sampled from $\mathcal{N} (0,\sigma^2)$,
\item rotation: choose $a$ uniformly at random in $\lbrace 1, 2, \ldots, 8 \rbrace$ and apply a rotation of $a \frac{\pi}{4}$ to the image,
\item ``junk \da'': replace the inputs with random images where each pixel is drawn uniformly in $\left[0,1 \right]$.
\end{itemize}
Obviously, the third \da should slow down the learning, while the remaining ones are expected to improve it.

\subsection{Experiments}

The tested Unet is implemented using Keras with Tensorflow backend and has the same architecture as \cite{ronneberger2015u} (8 layers, $3\times 3$ filters + ReLU). However, we use zero-padding so that output shapes match input ones. 
The layers contain (from 1st to last)  $10, 20, 40, 80, 80, 40, 20$ and $10$ filters respectively. 
Except for the last layer, we use an $\mathcal{L}_2$ regularization penalty with regularization factor $1e-4$. 
The neural network optimizer is RMSProp with learning rate $1e-4$ (other parameters are left to default Keras values) which minimizes quadratic train-loss $L \left( f_{\boldsymbol\theta} \left( \mathbf{x} \right) ,y \right) = \frac{1}{n} \sum_{\left( \mathbf{x},y \right)\in \mathcal{D}  } \left\lVert f_{\boldsymbol\theta} \left( \mathbf{x} \right) - y\right\rVert_2^2$. 

To assess the performances of our \da allocation algorithm, we train \unet 10 times and report average Jaccard index computed on a test set (disjoint from the training set) containing $30$ samples. The Jaccard index is the number of pixels in the intersection of the predicted mask region and ground-truth ($y$) divided by the number of pixels in the union of the predicted and true masks. The predicted mask is obtained by comparing $f_{\boldsymbol\theta}(\mathbf{x})$ to a threshold of $0.5$. 
The number of augmented data per epoch is $n_a=60$.  We compare SODA with 2 concurrent strategies: a uniform allocation over the 3 \da types and a ``target'' allocation which ignores the junk \da and evenly queries the remaining two. Ideally, SODA should be able to automatically discard the junk \da and achieve comparable performances without any prior knowledge on the usefulness of each \da type.

\begin{figure}[!h]
     \centering
         \centering
         \includegraphics[width=0.87\linewidth]{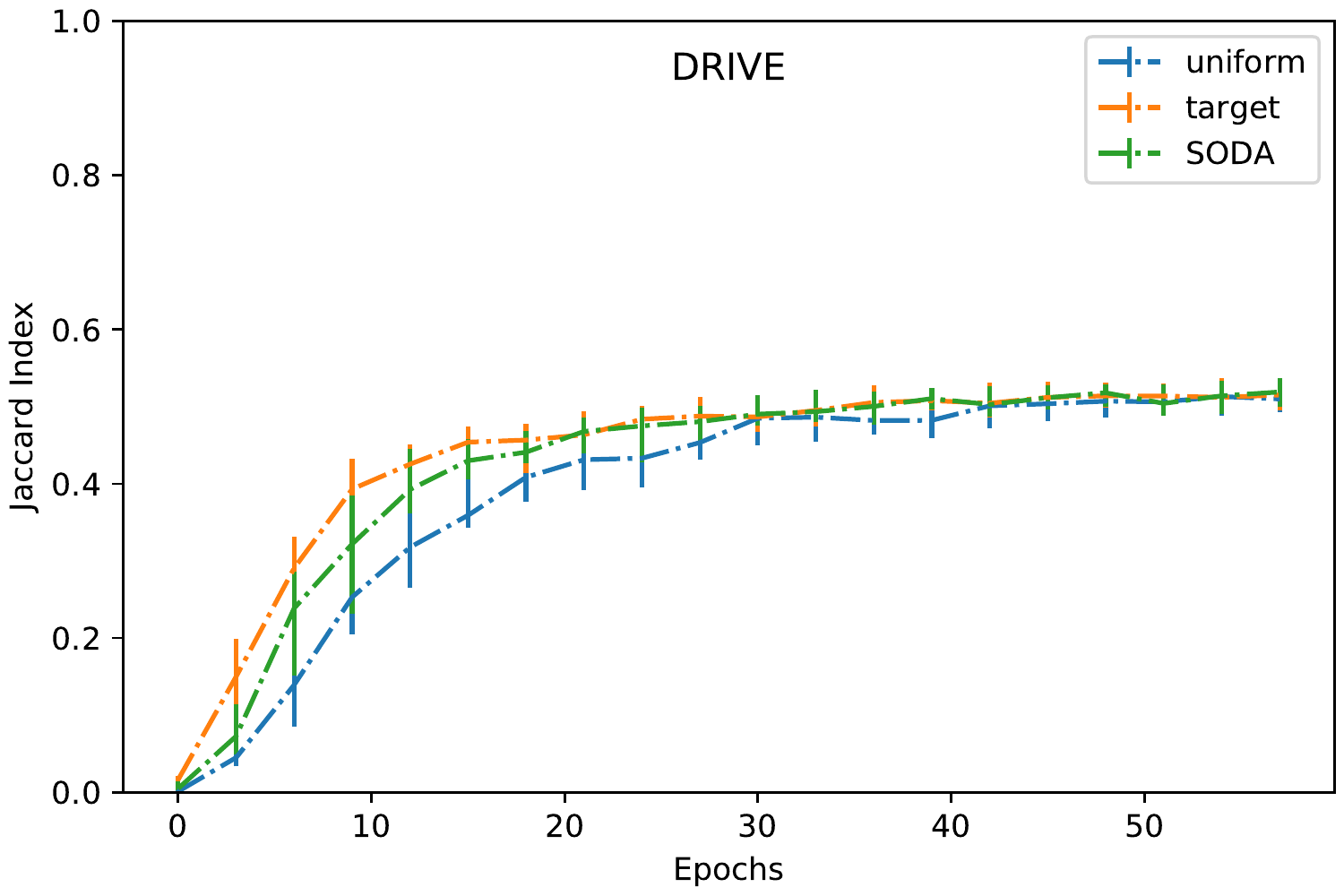}
         
         \includegraphics[width=0.87\linewidth]{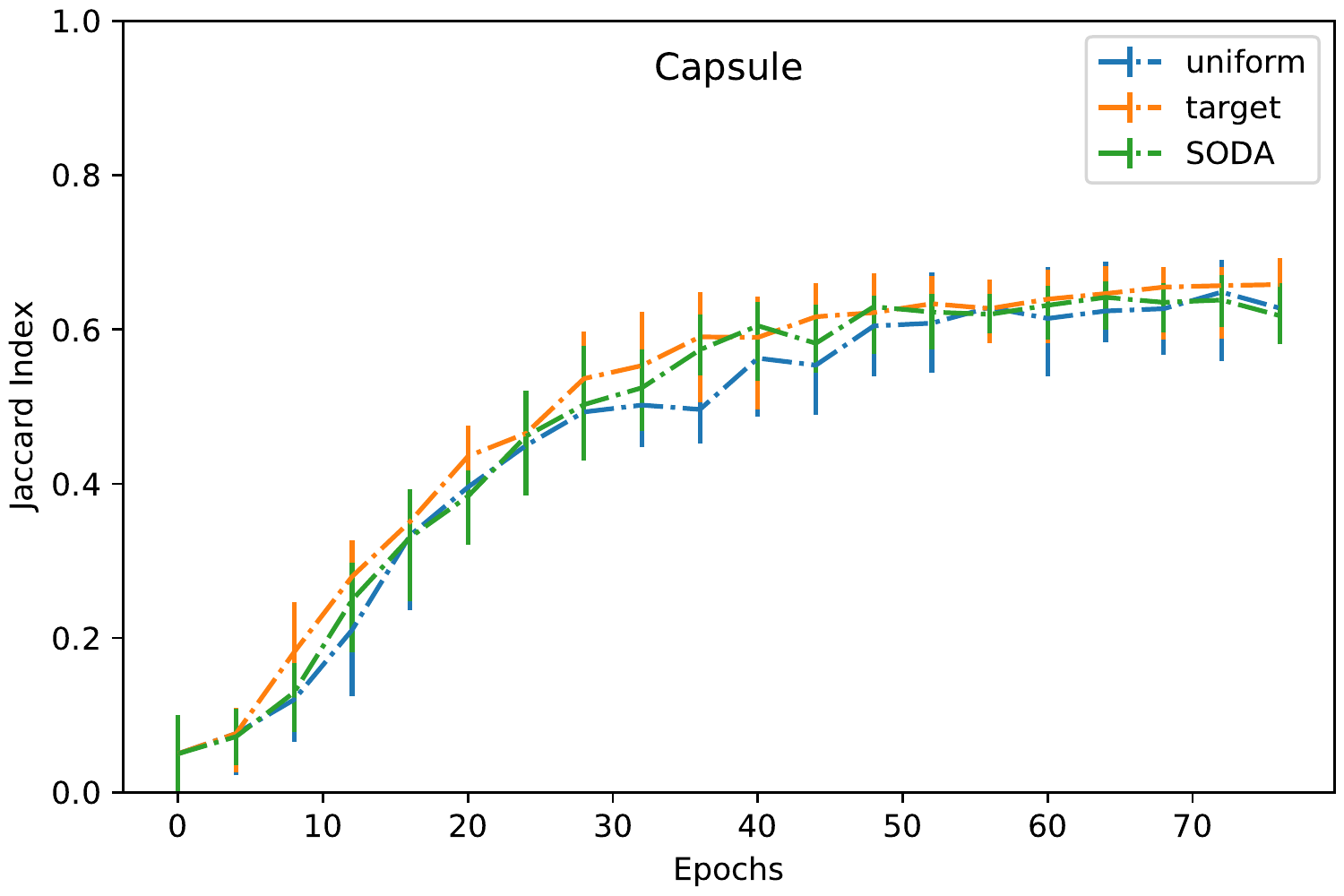}
         
         \includegraphics[width=0.87\linewidth]{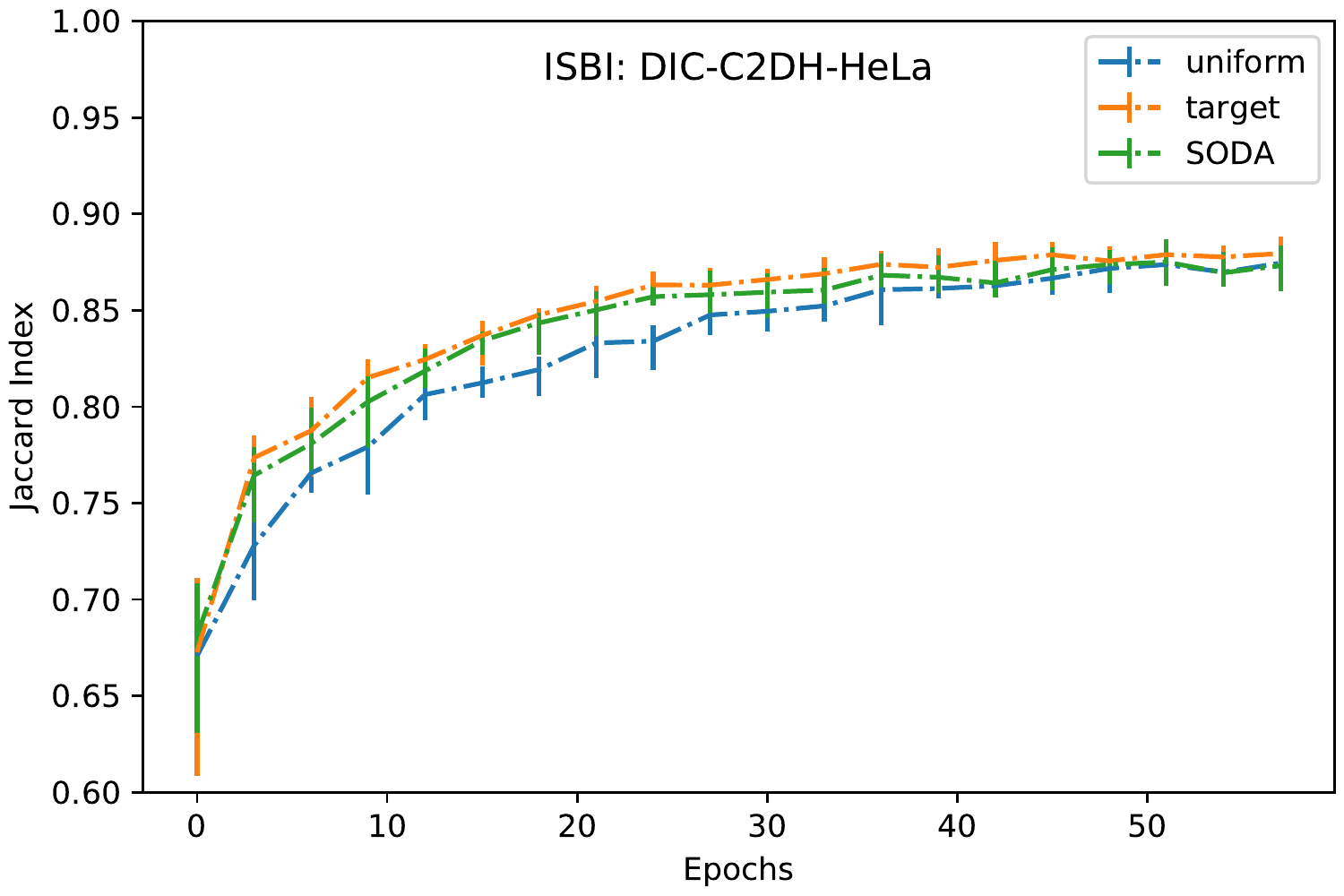}
         
         \includegraphics[width=0.87\linewidth]{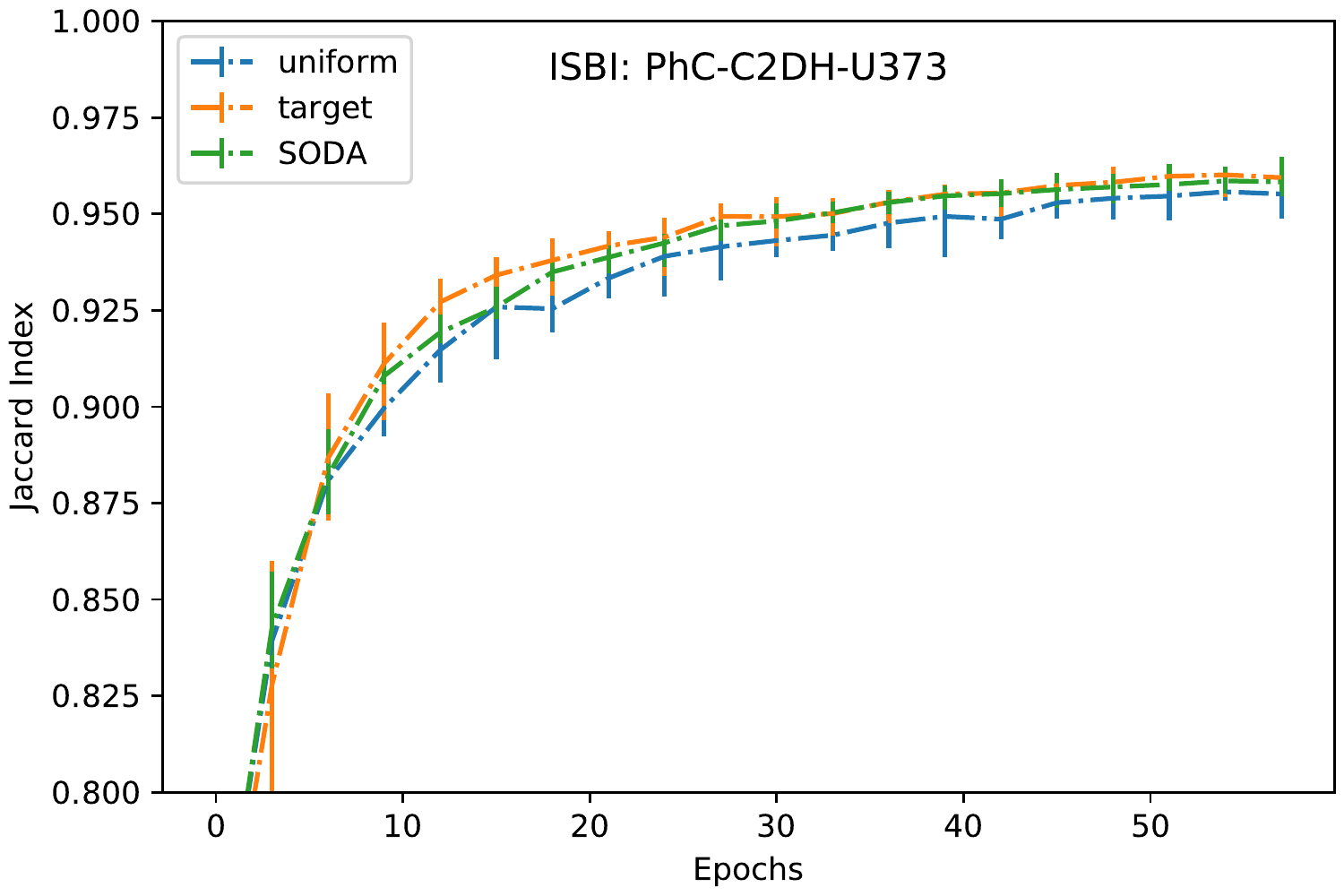}
         
\vspace{-1em}
\caption{Accuracy w.r.t to epochs for the 4 datasets and the 3 allocation strategies.\label{fig:acc}}
\end{figure}

For each dataset, Table \ref{tab:params} gives SODA parameters, while Figure \ref{fig:acc} shows the Jaccard index over epochs achieved by the different strategies. We can see that SODA outperforms the uniform allocation, regardless of the dataset. Its performances are also very close to the target strategy. In the last epochs, all strategies generally converge to close Jaccard index values. 
SODA learns at a faster pace which is useful in strongly computationally constrained learning environments where only a small number of epochs are possible.

\begin{table}[!h]
\begin{center}
 \begin{tabular}{ l | c | c | c }
    Dataset & $\eta$ & $\rho$ & $\beta$ \\
 \hline
 \hline
   DRIVE database & 6 & 0.99 & 0.5 \\
   Capsule & 4 & 0.99 & 0.5 \\
   ISBI: DIC-C2DH-HeLa & 3 & 0.99 & 0.5 \\
   ISBI: PhC-C2DH-U373  & 7 & 0.99 & 0.5 \\
 \hline  
 \end{tabular}
 \end{center}
 \vspace{-0.5em}
 \caption{SODA hyperparameters for each dataset. We only vary $\eta$ and keep $\rho$ and $\beta$ to the same value for all datasets which shows that hyperparemeter setting can essentially focus on SODA learning rate. \label{tab:params}}
\end{table}

\section{Conclusion} 
\label{sec:conclusion}

This paper introduces SODA, an algorithm for Self-Organizing Data Augmentation that can be deployed within usual deep neural network training loops. This algorithm leverages online learning to identify relevant allocation of computational resources for each type of data augmentation.
%
We show that the benefits of each type of data augmentation can be monitored through the dot product of two average gradients: the one from the data augmentation itself, and the one obtained from data points belonging to the original training set.
To account for the time variability of meaningful allocations, a discounted version of the algorithm is used.

In our experiments on segmentation in biomedical images using a \unet architecture, SODA proves to be able to outperform a standard uniform distribution of computational resources with respect to data augmentation types. According to our results, SODA achieves comparable accuracy w.r.t. the uniform policy but at a faster rate, which may be helpful for reducing ML carbon impact. 

In future works, we wish to generalize SODA to situations where there is no relevant data augmentation to choose from. While SODA exhibits improved robustness, we also believe the action-loss signals could be further processed in order to discard irrelevant data augmentation types more easily.

\vspace{1em}
\noindent \textbf{Acknowledgements}: This work was performed using HPC resources from GENCI–IDRIS (Grant 2021-AD011011606R1).



\bibliographystyle{IEEEbib}
\bibliography{ref}

\end{document}